\documentclass[conference]{IAC}

\usepackage{amsmath,amssymb,amsfonts}
\usepackage{algorithmic}
\usepackage{graphicx}
\usepackage{textcomp}
\usepackage[dvipsnames]{xcolor}
\usepackage{hyperref}
\usepackage[absolute]{textpos}
\usepackage{tabu}
\usepackage{textcomp}
\usepackage{titlesec}
\usepackage[colorinlistoftodos]{todonotes}
\usepackage{svg}
\usepackage{subcaption}

\titleformat{\subsection}[hang]
{\normalfont\normalsize\itshape}
{\thesubsection}
{1em}
{#1}

\usepackage{lastpage}
\usepackage{fancyhdr}
\usepackage{lipsum} 
\usepackage{xcolor}

\usepackage{makecell}

\usepackage[style=numeric-comp,sorting=none]{biblatex}

\def\BibTeX{{\rm B\kern-.05em{\sc i\kern-.025em b}\kern-.08em
    T\kern-.1667em\lower.7ex\hbox{E}\kern-.125emX}}

\usepackage[nolist,nohyperlinks]{acronym}
\acrodef{mlt}[MLTs]{Martian Lava Tubes}

\begin{document}

\begin{textblock}{6}(5,1.9)
\centering
\noindent \large  IAC-23-D3.IPB.1
\end{textblock}

\title{\vspace{2ex} Martian Lava Tube Exploration Using Jumping Legged Robots:\\ A
Concept Study \vspace{-2ex} 

{\footnotesize \textsuperscript{  } }
}

\author{ \large\textbf{Jørgen Anker Olsen}$^{a,1}$, \large\textbf{Kostas Alexis}$^{b,}$}

\affil[ ]{\color{gray} $^1$ Department of Engineering Cybernetics, Norwegian University of Science and Technology, Trondheim, Norway}

\affil[a]{jorgen.a.olsen@ntnu.no,  $^b$ konstantinos.alexis@ntnu.no}

\twocolumn[
  \begin{@twocolumnfalse}

\maketitle

\begin{abstract}
\centerline{\large\textbf{Abstract}}
In recent years, robotic exploration has become increasingly important in planetary exploration. One area of particular interest for exploration is Martian lava tubes, which have several distinct features of interest. First, it is theorized that they contain more easily accessible resources such as water ice, needed for in-situ utilization on Mars. Second, lava tubes of significant size can provide radiation and impact shelter for possible future human missions to Mars. Third, lava tubes may offer a protected and preserved view into Mars' geological and possible biological past. 

However, exploration of these lava tubes poses significant challenges due to their sheer size, geometric complexity, uneven terrain, steep slopes, collapsed sections, significant obstacles, and unstable surfaces. Such challenges may hinder traditional wheeled rover exploration. To overcome these challenges, legged robots and particularly jumping systems have been proposed as potential solutions. 

Jumping legged robots utilize legs to both walk and jump. This allows them to traverse uneven terrain and steep slopes more easily compared to wheeled or tracked systems. In the context of Martian lava tube exploration, jumping legged robots would be particularly useful due to their ability to jump over big boulders, gaps, and obstacles, as well as to descend and climb steep slopes. This would allow them to explore and map such caves, and possibly collect samples from areas that may otherwise be inaccessible.

Traditional terrestrial legged robots such as ANYbotics ANYmal and Boston Dynamics Spot have reached the commercial market and proven their capability and robustness, while both had a prominent role during the DARPA Subterranean Challenge. This together with the work done with NASA's Scorpio, ETH's SpaceBok, and others has opened the possibility for legged systems with their unique designs and capabilities to be used in Martian lava tube exploration.

This paper presents the specifications, design, capabilities, and possible mission profiles for state-of-the-art legged robots tailored to space exploration, focusing on jumping and walking for locomotion, in-air stabilization, and jumping trajectories. Additionally, it presents the design, capabilities, and possible mission profiles of a new jumping legged robot for Martian lava tube exploration that is being developed at the Norwegian University of Science and Technology. 

\end{abstract}

  \end{@twocolumnfalse}
]

\section{Introduction}

Robotic exploration has been pivotal for remote study of neighboring planets, yielding significant scientific findings. These insights encompass Mars' geological history, geochemical composition, and the presence of surface water ice, thus underpinning strategies for future human missions~\cite{water_mars2022}.

While the selection of exploration sites aligns with specific mission objectives and robotic capabilities, most endeavors have historically favored the safer, flatter terrains of Mars. Previous and current missions on Mars consist of a singular lander, lander and rover, rover only~\cite{bell2000mineralogic, arvidson2011opportunity,grotzinger2012mars} and newly a rover and helicopter combination~\cite{balaram2021ingenuity}.

After the identification of Martian lava tubes from space in the early 1970s~\cite{carr1973volcanism}, these geological cavities have garnered significant scientific interest due to their potential for revealing Mars' subsurface geology, giving insight into the possible biological past on Mars, and their potential for offering sheltered habitats for future human missions~\cite{boston2001cave,mari2021potential}. 

Lava tubes are created when a region with high volcanic activity, often near volcanoes, is filled with surface or subsurface lava channels where the outer surface of the channel that is transporting lava cools more rapidly than the rest of the lava flow, creating a hardened crust. This is subsequently followed by the lava inside the tube flowing out, leaving an empty void and creating the lava tube~\cite{leveille2010lava}. Several variants of lava tubes exist, depending on their lava flow, type of solidification, closeness to the surface, and erosion by the lava flow. The size of these lava tubes can reach up to $400$ meters in diameter~\cite{sauro_lava_2020}. These tubes are numerous and spread out in basaltic volcanic surface structures along the Martian surface like collapsed chains or rilles. Collapsed rilles along with single circular collapsed roof segments of the lava tubes, called skylights, are the most commonly proposed entry points of these underground cavities. Although there have not been any close-up investigations of these possible entry points, satellite imagery and lava tube entrances on Earth indicate that they have been created by the partial or full collapse of the lava tube roof and are thought to leave behind a significant pile of rubble and boulders. These lava tube entrances, through a collapsed rille entry or a vertical skylight, may prove impossible for the existing exploration systems to overcome, necessitating novel systems and approaches~\cite{mari2021potential}. 

NASA's recent mission successfully demonstrated the use of a flying drone, the Mars helicopter Ingenuity, as an exploration platform~\cite{balaram2021ingenuity}. However, its low payload capabilities and short flight time limit its versatility underground in the large-scale and enclosed environment of a Martian lava tube. Moreover, the perceptually degraded lave tube environment, especially due to the presence of Martian dust, and complex walls and boulders structures can prove fatal to a helicopter, especially in case of errors such as localization drift leading to a catastrophic collision.

Several specialized and innovative solutions have emerged in pursuing advanced robotic explorations within Martian lava tubes. Noteworthy examples tailored for lava tube exploration include ReachBot~\cite{newdick2023designing}, DEADELOUS~\cite{rossi2021daedalus}, and PitBot~\cite{thangavelautham2017flying}. Additionally, versatile systems suitable for a broader range of terrains encompass legged, jumping, or climbing robots such as SpaceBok~\cite{arm_spacebok_2019}, ANYmal~\cite{arm2023scientific}, Space Climber~\cite{bartsch2010spaceclimber}, CLOVER Robot~\cite{macario2022clover}, and NASA Athlete~\cite{wilcox2007athlete}. As we consider the unique challenges presented by steep and vertical entry points like skylights, deploying systems within these formations necessitates specialized mechanisms. RoboCrane~\cite{miaja2022robocrane} serves as a support example, offering a crane-based solution designed to safely lower robots into such demanding environments.

In this work, we will outline a concept for Martian lava tube exploration using legged robots capable of high jumping maneuvers that can prove essential to overcome tall obstacles or pass over ditches. We will first overview other concepts for Martian Lava Tube exploration, present the case for legged robots, before diving deeper into the specific case and potential of jumping legged robotic systems. We will further present our own quadrupedal jumping legged robot concept, designed to traverse normal Martian terrain by walking and utilizing jumping to overcome significant obstacles expected to be found at the entrance and inside Martian lava tubes. Importantly, the robot has the ability to reorient itself in-flight before landing by utilizing its legs to perform in-flight stabilization, which is crucial due to the initial spin from a powerful needs to be canceled out to ensure a safe landing.

Section~\ref{sec:related} describes in detail the Martian environment, terrain, and lava tubes. In Section~\ref{sec:RoboticExploration}, we survey historical and contemporary robotic exploration methods, highlighting their specific capabilities. Section~\ref{sec:Novel} underscores the rationale behind employing specialized solutions for targeted exploration activities. In Section~\ref{sec:050_JumpingLeggedSystems} a selection of legged robots is presented. Our proposed jumping quadrupedal robot design is presented in Section~\ref{sec:Designs}. Simulations based on this design are examined in Section~\ref{sec:070_Simulation}. In Section~\ref{sec:080_MissionProfile}, we describe a possible mission profile for the proposed jumping quadrupedal robot system.  While Section~\ref{sec:conclusions} offers concluding remarks and future work.

\section{Martian Environment, Terrain \& Lava Tubes}\label{sec:related}

This section overviews the Martian environment and terrain, alongside lava tubes and their significance, challenges, and possible entry points.

\subsection{Martian atmosphere, surface, and terrain}
Mars is often considered to be the most viable target for human and robotic exploration in our solar system due to its proximity to Earth. This, together with indications of past liquid water, similar day length, polar ice caps, and its relative similarity in size, underlines its importance for human exploration~\cite{barlow_2008}. The Martian atmosphere composition is dominated by Carbon dioxide (CO2) at $95.32\%$  and only contains $0.13\%$  oxygen, and the atmospheric pressure on the surface is $670$ Pascal, less than $1\%$  that of Earth~\cite{barlow_2008}. The temperature of the surface of Mars contributes to making it an inhospitable place, with a mean surface temperature around the equator of $215$ Kelvin and very large fluctuations between day and night temperatures due to the thin atmosphere \cite{leveille_lava_2010}.

Mars has a gravity of $3.72\ \frac{m}{s^2}$ or $37.9$ \% of Earth's gravitational force. This diminished gravitational pull results from its smaller size and mass compared to Earth. The Martian gravity directly influences various aspects of exploration, such as engineering requirements for landers, rovers, and other surface exploration systems. This necessitates specialized designs and considerations for effective and safe missions on the Martian surface~\cite{hirt2012Gravity}.

Mars' surface is constantly subjected to significant space radiation, primarily due to two distinct factors: the absence of a meaningful magnetosphere and its exceedingly thin atmosphere. Unlike Earth's magnetosphere, the planet's lack of a protective magnetic field makes it vulnerable to cosmic and solar radiation. The thin atmospheric composition, dominated by carbon dioxide, also offers minimal shielding against radiation influx, posing challenges for potential manned missions and surface systems durability~\cite{hassler2014mars,paris2019prospective}

\subsection{Martian lava tubes}

Owing to ancient volcanism and Mars' minimal erosion, geological formations like volcanoes, valleys, river deltas, and lava tubes from its distant active era billions of years ago remain well-preserved. This preservation offers a unique window into the planet's historical geophysical processes and climatic conditions. The Martian terrain has been extensively mapped through high-resolution multi-spectral imaging from orbiters and surface-based rovers. However, little is known about the subsurface of Mars and its potential for underground caves and tunnels~\cite{sauro_lava_2020}.  

From satellite imagery of Mars, many possible collapsed and not collapsed lava tubes and caves have been observed. These have been compared with similar lava tubes on Earth, resulting in very high confidence in their presence on Mars and giving indications of the ranges of sizes the lava tubes on Mars can achieve compared to those on Earth~\cite{leveille2010lava,sauro_lava_2020}. 

Lava tubes are defined as ``roofed conduit of flowing lava, either active, drained, or plugged''~\cite{gunn2004encyclopedia}. Lava tube widths on Earth vary from $0.5$ to $30$ m, while on Mars, the width can reach up to $400$ m. This gives the Martian lava tubes $1-3$ orders of magnitude more volume than on Earth. These vast dimensions and their closeness to the surface, at a distance of a few tens of centimeters to a few tens of meters, suggest that Martian lava tubes could serve as potential habitats~\cite{sauro_lava_2020}. 

Exploring subsurface lava tubes on Mars is crucial for several reasons. Firstly, it provides a unique opportunity to study the planet's geological past by accessing non-eroded rock formations. Secondly, it could lead to the discovery of valuable resources like water-ice and minerals that could be used for future missions. Thirdly, these underground spaces could serve as a potential habitat for human exploration and colonization. Lastly, exploring these underground environments could also reveal clues about the possible biological past of Mars. Additionally, the inside of the cave offers shelter from the Martian weather, dust storms, radiation, micro-meteorites, and has a more stable internal environment and temperature, protecting from the heat cycles caused by the day-night cycle~\cite{mari2021potential}.

\begin{figure}[t]
    \centering
    \begin{subfigure}[b]{0.49\textwidth}
        \centering
        \includegraphics[width=\textwidth]{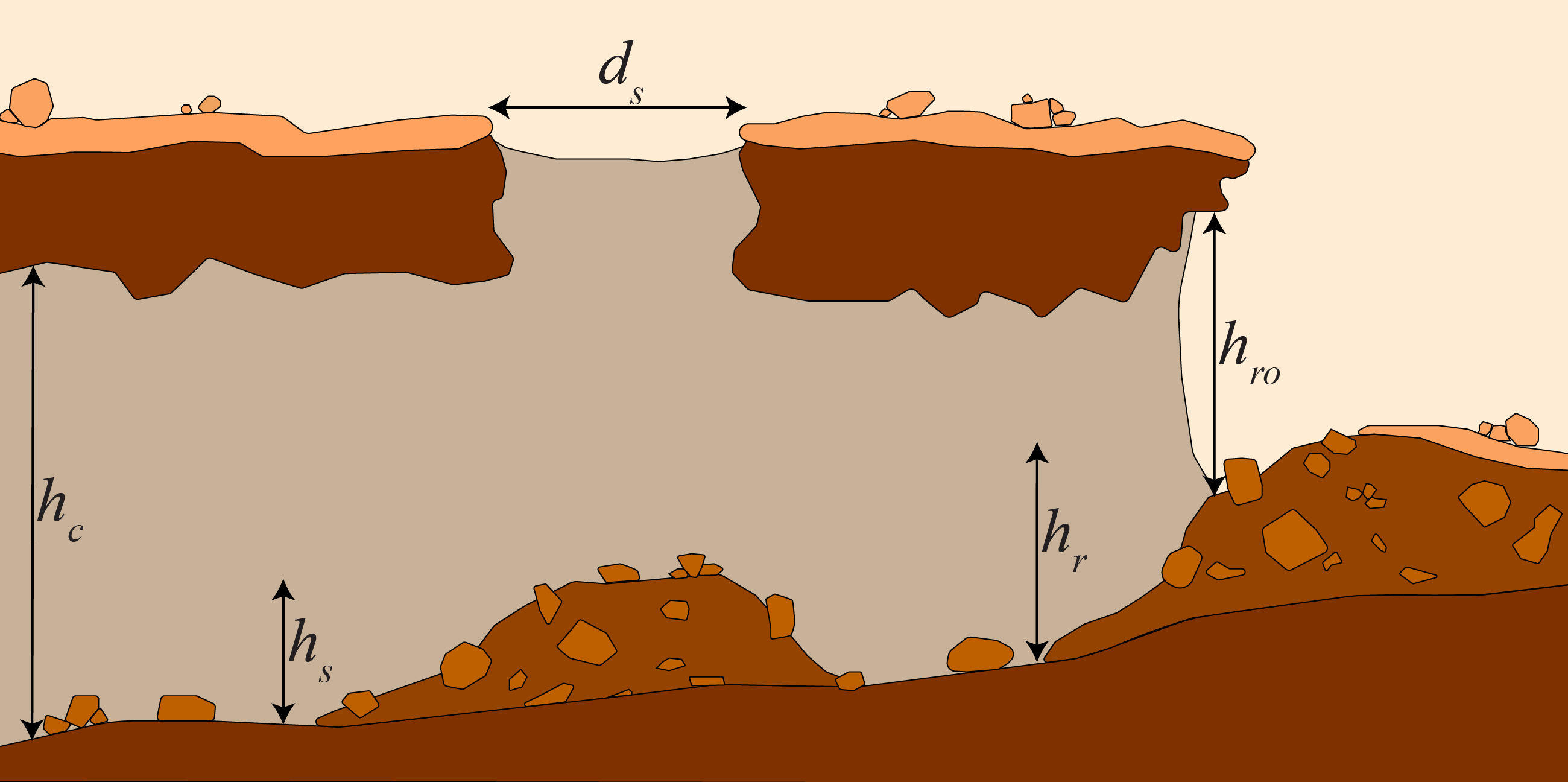}
        \caption{Side view representation of a Martian lava tube with skylight and rille entrance.}
        \label{fig:first_sub}
    \end{subfigure}
    \begin{subfigure}[b]{0.49\textwidth}
        \centering
        \includegraphics[width=\textwidth]{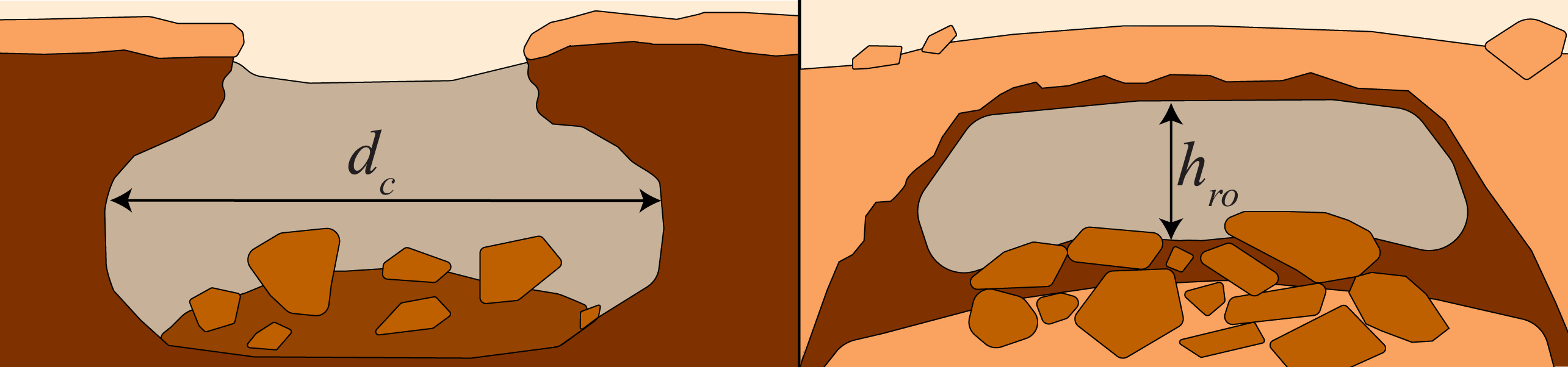}
        \caption{Front view of a Martian lava tube, on the left skylight, on the right rille entrance.}
        \label{fig:second_sub}
    \end{subfigure}
    \vspace{-4ex}
    \caption{Martian lava tube depiction: (a) presents a longitudinal perspective of a Martian lava tube featuring a skylight opening and an associated rubble accumulation. It also indicates a potential rille entrance and its related rubble. In (b), the left diagram provides a cross-sectional view of the skylight and the tube's internal diameter, whereas the right displays the rille entrance's frontal view. $h_{c}$ is the height of the cave, $d_c$  is the horizontal diameter of the cave, $d_s$ is the diameter of the skylight opening, $h_s$ is the height of the skylight rubble pile, $h_r$ is the height of the rille rubble pile, $h_{ro}$ is the rille opening height.  }
    \label{fig:lavaTubeCrossSection}
\end{figure}

\subsection{Access to Martian lava tubes}

There is limited information about the structure of lava tubes on planets other than Earth since they have not been directly investigated. The available data comes from satellite imagery, modeling, and Earth analog caves. To enter uncollapsed sections of Martian lava tubes, one can use the natural outflow opening of the lava, a rille where the cave could lead into the lava tube, a skylight of the lava tube, or by drilling or blasting an opening through a section of the lava tube that is close enough to the surface~\cite{paris2019prospective}.

Robotic exploration of Martian lava tubes faces challenges due to their access points. Rilles and horizontal openings often lead to substantial accumulations of debris, ranging from boulders to rubble, a consequence of partial or full cave collapses. Skylights, formed due to the caving-in of lava tube roofs, pose similar issues with their boulders, rubble, or sand accumulations that could hinder robotic traversal~\cite{sauro_lava_2020}. Vertical descents are required for skylights and artificially drilled entry points, which necessitates specialized deployment equipment like cranes, cables, or lowering mechanisms to facilitate the robot's descent, followed by intricate maneuvering to navigate through the debris zone~\cite{miaja2022robocrane}. Fig.~\ref{fig:lavaTubeCrossSection} illustrates a longitudinal view of a lava tube along with possible skylight and rille entrances.



\section{Robotic Exploration: Capabilities and Limitations of Current Systems}\label{sec:RoboticExploration}

Having outlined the significance and challenges of exploring Martian subsurface caverns and lava tubes, it becomes evident that traditional methods may not be suitable for such complex and challenging environments. The unique demands of these lava tubes necessitate a comprehensive understanding of current robotic capabilities. This section delves into the robotic systems and other solutions used so far in Martian exploration, highlighting both their strengths and limitations in the context of extraterrestrial exploration.

\subsection{Orbiters \& Landers}

Over the years, space agencies have launched a number of Mars orbiters that have been vital in expanding our knowledge of the Red Planet. These spacecraft are equipped with advanced cameras and scientific instruments that collect crucial data on Mars' atmosphere, climate, topography, and geology. They capture detailed images of the Martian surface, which helps identify suitable landing sites for rovers and potential future manned missions. Additionally, they act as communication relays between Earth and the rovers on Mars, ensuring uninterrupted data transmission.

Thanks to observations by these orbiters, scientists have gained a better understanding of the history of Mars, its present conditions, and the potential for future exploration. Some of the noteworthy past orbiters include Mars Odyssey~\cite{saunders20042001}, Mars Express~\cite{chicarro2004mars}, Mars Reconnaissance Orbiter (MRO)~\cite{malin2007context}, Mars Global Surveyor (MGS)~\cite{albee2001overview}, ExoMars Trace Gas Orbiter (TGO)~\cite{thomas2017colour}, MAVEN (Mars Atmosphere and Volatile EvolutioN)~\cite{jakosky2015mars}, and Tianwen-1~\cite{zou2021scientific}

On the surface of Mars, the Viking 1 and Viking 2 landers were pioneers in Martian exploration in the 1970s. They provided initial meteorological insights~\cite{hess1977meteorological}. The Phoenix lander discovered water-ice presence, while InSight offered profound geological observations that deepened our knowledge of Martian terrains and subsurface structures~\cite{golombek2018geology}.

\subsection{Rovers}
The campaigns for Martian rovers started with the lander Mars Pathfinder and its accompanying rover Sojourner~\cite{bell2000mineralogic}. In the following decades, numerous rovers started roving the Martian surface. They all utilize a similar form factor, the six-wheeled all-wheel drive. Some relatively small and less equipped, like Spirit~\cite{morris2010identification}, Opportunity~\cite{arvidson2011opportunity}, compared to the recent larger and much more capable rovers of Curiosity~\cite{grotzinger2012mars}, Perseverance~\cite{mangold2021perseverance}, and  Tianwen-1~\cite{zou2021scientific}

Future planned rovers, such as ExoMars Rosalind Franklin rover~\cite{quantin2021oxia}, also follow similar design characteristics as the previously mentioned rovers, while the  Mars Sample Return(MSR) rover may differ slightly with four wheels due to its unique mission~\cite{tait2022preliminary}.

These rovers were created to conduct exploration and scientific tasks in relatively even terrain. If the terrain becomes too challenging, the current rovers may encounter difficulties. The benefits of using wheeled locomotion include the fact that it is a well-established and widely utilized technology on Mars, as well as its high energy efficiency, stability, substantial payload capacity, and redundancy in the event of wheel failure. There are numerous designs available to enhance their traversing capabilities, taking into account differences in size, wheel suspension, center of gravity, and ground clearance~\cite{seeni2008robot}.

Although very successful, some disadvantages of rovers are their tendency to get wheels slippage or stuck in the Martian sand and the need to maneuver around and avoid steep and rough terrain while also avoiding large obstacles~\cite{seeni2008robot}. 

\subsection{Helicopters}

The Mars helicopter Ingenuity, weighing in at $1.8$ kg, has rotor blades measuring $1.21$ m in diameter and a coaxial configuration. Its purpose was only to demonstrate the possibility of flight in the Martian atmosphere~\cite{balaram2021ingenuity}. Despite this, Ingenuity has surpassed expectations by completing over a dozen flights, reaching speeds of $5 \ \frac{m}{s}$  and altitudes of up to $12$ m. With each flight covering up to $600$ m, it has accumulated multiple kilometers of distance and proven the potential of flying robots on the surface of Mars. The Mars Science Helicopter, a larger version, is currently in development~\cite{withrow2020recent}.

The Mars helicopter has a significant advantage in its ability to fly and navigate over different types of terrain with relative ease. However, it also has its disadvantages, such as its limited flight time and the need for multiple days of sun exposure to recharge its batteries. Additionally, it requires flat terrain for landing and can be fragile due to its fast-moving rotors and sensitive electronics, which could be easily damaged even by a minor crash. The helicopter's payload capabilities are restricted because it must be extremely lightweight to fly in the less dense atmosphere of Mars. Apart from an engineering and landing camera, it is very limited in scientific payload capability.

\vspace{-2ex}
\section{Novel Martian Exploration Systems}\label{sec:Novel}
This section will cover a variety of robot concepts proposed for exploring Martian lava tubes, both as independent units or as part of a larger mission. We will also showcase other advanced robotic systems that were initially designed for extraterrestrial exploration in various contexts but could also be effective for exploring Martian lava tubes. The following subsection will focus on non-legged robots, while the next section focuses on legged systems. 

\subsection{Novel concepts for Lava Tube Exploration}

The nature of exploring and operating in an unknown environment, such as a Martian lava tube, necessitates certain critical capabilities: 1. Advanced sensing and mapping, 2. Autonomous navigation, 3. Decision making, 4. Energy efficiency. Martian lava tubes present many challenges for robotic systems and their deployment. The most notable challenges include traversing the Martian regolith, climbing steep slopes, and overcoming boulders or crevasses. Various robotic solutions have been proposed to overcome these challenges and to explore Martian lava tubes. 

As the lava tubes provide large walls and roofs, climbing as a form of locomotion and exploration can be advantageous as it gives a way to descend into a skylight and have an unobstructed overview of the lava tube while traversing the cave roof or walls.

ReachBot~\cite{newdick2023designing} and LEMUR 3~\cite{parness2017lemur} utilize this in their concepts. ReachBot is a  climbing cable-driven robot composed of a central body containing sensing, power, and computing. It utilizes cable-driven extendable booms with shoulder and wrist joints along with an end effector to climb the walls and roof, or traverse the floor in the lava tube. ReachBot deploys a varying number of booms based on the specific mission, modifying the workspace, stability, and weight of the robot. 
LEMUR 3 consists of a central body and four highly dexterous 7-degree-of-freedom arms, each equipped with a specialized gripper at the end to facilitate climbing~\cite{parness2017lemur}. To climb on rocky surfaces, it utilizes microspine grippers to attach itself to the rock. While being tethered, LEMUR 3 was able to traverse at a speed of $0.16\ \textrm{m/h}$. Since then, its speed has been substantially increased.

Robots leveraging a spherical form factor have also been proposed, such as Pit-bots \cite{thangavelautham2017flying} and DAEDALUS \cite{rossi2021daedalus}. Pit-bots, weighing $3 \ \textrm{kg}$ and measuring $0.3\ \textrm{m}$ in diameter, are engineered to fly, hop, and roll using onboard miniature thrusters for lunar and Martian lava tube exploration. Deployed as a trio, these robots employ thrusters and stereo cameras to navigate and map the lava tubes.
DAEDALUS is designed for tethered descent into lava tube skylights while using its sensors for mapping and imaging. Its outer shell incorporates a pole mechanism for movement, while its core houses rolling apparatus and electronics. Within the lava tube, DAEDALUS uses its spherical design for rolling or employs actuated poles to overcome obstacles it can not easily roll over.

Tethered wheeled rovers such as Moon Diver/Axel \cite{nesnas2012axel,nesnas2023moon} and Coyote III \cite{sonsalla2022towards} are engineered to cooperate with a rover or lander, facilitating the wheeled rover's descent into a lava tube via a skylight entrance. The Moon Diver's Axel configuration employs a bi-wheeled Axle tethered directly to an adjacent lander at the skylight opening. This direct connection offers the advantage of lander-provided power during descent, enabling a gradual, controlled descent with human oversight for operational decisions. On the other hand, Coyote III introduces a hybrid approach, utilizing wheeled-legged hybrid locomotion, meaning the wheels consist of multiple rigid legs. Integrated into a multi-robotic team, it pairs with a larger wheeled rover, which is equipped with a robotic arm responsible for tether management.

Jumping robots, capable of leaping multiple times their height, provide a distinct advantage in overcoming substantial obstacles. The CLOVER Robot \cite{lo2021energetic,macario2022clover} exemplifies this approach with its minimalist design, emphasizing maximal jump height while reducing potential failure points. This robot integrates small motors, rotational springs, and elastic components for its locomotion. When deployed in swarms they have potential applications in lava tube exploration.

The Cavehopper, referenced in \cite{whittaker2012technologies}, is a wheeled jumping robot designed for dual functionality. It combines the capabilities of a compact rover with the ability to leap, facilitating both entry into and navigation within lava tubes. Intended for swarm deployment, it operates alongside a lander equipped with a tethered communication node. Its robust design allows the Cavehopper to enter lava tube skylights through a direct jump, withstanding rugged landings. Each unit is equipped with miniaturized scientific instruments.

\subsection{Support systems}
The previously described robotic systems primarily emphasize exploration capabilities and some rely on supporting rovers or landers. RoboCrane \cite{miaja2022robocrane} is conceived as a support system that can lower an exploration robot into a skylight entrance. Subsequent to its release, RoboCrane facilitates power and establishes a communication link between the lava tube and the planetary surface. Such support is crucial for many proposed missions, given that robots tailored for lava tube traversal often prioritize terrain adaptability over extended operations and Earth-bound communication.

The development of robotic systems for Mars encounters distinct challenges and limitations. Many existing robotic concepts focus on specific elements of lava tube exploration but frequently lack the versatility and advanced mobility needed for navigating various obstacles and terrains, areas where legged robots might excel.

\vspace{-2ex}
\section{Jumping and legged systems }\label{sec:050_JumpingLeggedSystems}

\subsection{Legged Robots}

Legged robots have demonstrated increasing proficiency in navigating challenging, rugged, and adverse terrains, and have proven adept at overcoming obstacles and dynamically adapting to changes in the ground environment they traverse~\cite{tranzatto_cerberus_2022}. In order to succeed in space exploration, it is crucial to possess these attributes.

Several terrestrial state-of-the-art legged robotic systems have been studied for general or specific uses on the moon and Mars. The most notable that utilize dynamic locomotion and have been proposed for space exploration are ANYbotics ANYmal~\cite{hutter_anymal_2016,arm2023scientific}, Boston Dynamics Spot~\cite{spot_website,spot_website_nasa}, and DARPAs Legged Squad Support System (LS3)~\cite{mari2021potential}. Serial linkages with 3 or more degrees of freedom per leg are used by all.

Dynamic locomotion can be challenging for quadrupedal robots, but ANYmal demonstrates the versatility of use in industrial settings, search and rescue, data collection, mapping, and exploration~\cite{tranzatto_cerberus_2022}. In~\cite{arm2023scientific}, the authors present a multi-robot team of three ANYmal robots with diverse payloads for lunar surface exploration, a concept that could be extended to Martian lava tube investigations. Specialized locomotion policies and locomotion techniques intended for planetary exploration have also been tested on a modified ANYmal robot, and demonstrated extreme traverse capabilities in step terrain, up to $45\ \textrm{deg}$ in simulation~\cite{weibel2023towards}.

Spot has showcased robust dynamic walking capabilities and has been employed for inspection, exploration, and mapping tasks across various challenging terrains~\cite{agha2021nebula}. NASA has explored Spot's potential for lava tube investigations, conducting tests in terrestrial lava tubes~\cite{morrell2022nebula}. 

In~\cite{mari2021potential}, the LS3 robot was identified as a suitable legged system for exploration and cargo transport. This selection positions LS3 to operate alongside humans during manned Martian lava tube exploration scenarios, underscoring its potential in such specialized terrains.

Other walking legged robot prototypes developed for space exploration, such as SpaceClimber formerly SCORPION~\cite{bartsch2010spaceclimber,dirk2007bio} and NASA Athlete~\cite{wilcox2007athlete}, can utilize static walking as a form of locomotion, three or more legs are utilized for a more more stable walking gate. While NASA's Athlete robot is equipped with wheels at the end of its legs for efficient traversal on flat and firm terrain.

\subsection{Jumping legged robots}
Certain locomotion strategies, including walking, become more energy-efficient in planetary environments with gravity lower than Earth's. This efficiency arises from the reduced power requirements of the actuators during robot movement from the lower gravity. Similar advantages apply to locomotion methods such as pronking and jumping. SpaceBok~\cite{arm_spacebok_2019} capitalizes on this, integrating motors and springs to facilitate a combination of jumps and static walking. However, jumping introduces complexities, primarily due to the loss of ground contact and inherent stability. An unintended rotation can be imparted on the robot during lift-off. To counteract this, SpaceBok incorporates a reaction wheel, stabilizing its rotational velocity and facilitating reorientation for landings~\cite{kolvenbach2019towards}. SpaceBok utilizes a 4-bar closed kinematic design for its legs, with two motors and two degrees of freedom per leg utilized for jumping and walking. 

Terrestrial walking robots are designed with open kinematic linkages optimized for dynamic walking. For jumping robots, closed kinematic chains offer advantageous performance in jumping, as exemplified by the quadrupedal Stanford Doggo's leg design~\cite{kau2019stanford}.

These legged robotic solutions exemplify the versatility and use of legged and jumping legged robots and their possible use for Martian lava tube exploration.

\vspace{-2ex}
\section{The OLYMPUS Jumping Legged Robot}\label{sec:Designs}
The conceptual design of the OLYMPUS jumping legged robot is presented below. 

\subsection{Design}
To effectively explore Martian lava tubes and to overcome the expected large obstacles and rough terrain when entering lava tubes through skylights or rille entrances, we present OLYMPUS, a jumping legged robot concept. The robot is designed towards the following requirements: 1. Optimized jump height,  2. Ability to perform controlled in-flight attitude maneuvers, 3. Ability for dynamic walking, and 4. Retain two energy-efficient resting postures.

To achieve these goals, a quadrupedal system was designed, where the major factor in achieving the aforementioned list of requirements is the jumping capability of the leg and its ability to be precisely controlled in a large workspace. With the major focus on the jump height and ability to land comes the design of the robot leg. Traditional terrestrial state-of-the-art legged systems such as ANYbotics ANYmal~\cite{hutter_anymal_2016}, Boston Dynamics Spot~\cite{spot_website}, and MIT mini cheetah~\cite{katz_mini_2019} utilize an open kinematic chain in the form of a serial linkage with three actuators allowing for 3 degrees of freedom, with actuators in the hip and knee. Another -- albeit less common-- leg design is the closed kinematic chain in the form of a 4 or 5-bar linkage, such as in Minotaur~\cite{kenneally_design_2016}, Stanford Doggo~\cite{kau2019stanford}, and SpaceBok~\cite{arm_spacebok_2019}. The leg design chosen for OLYMPUS is a 5-bar design that offers several benefits, including an significant range of motion, large workspace, and greater efficiency in motor usage during a jump~\cite{kenneally_design_2016,kenneally_leg_2015,dong2017design, kau2019stanford}.

\begin{figure}[t]
    \centering
    \includegraphics[width =0.44\textwidth]{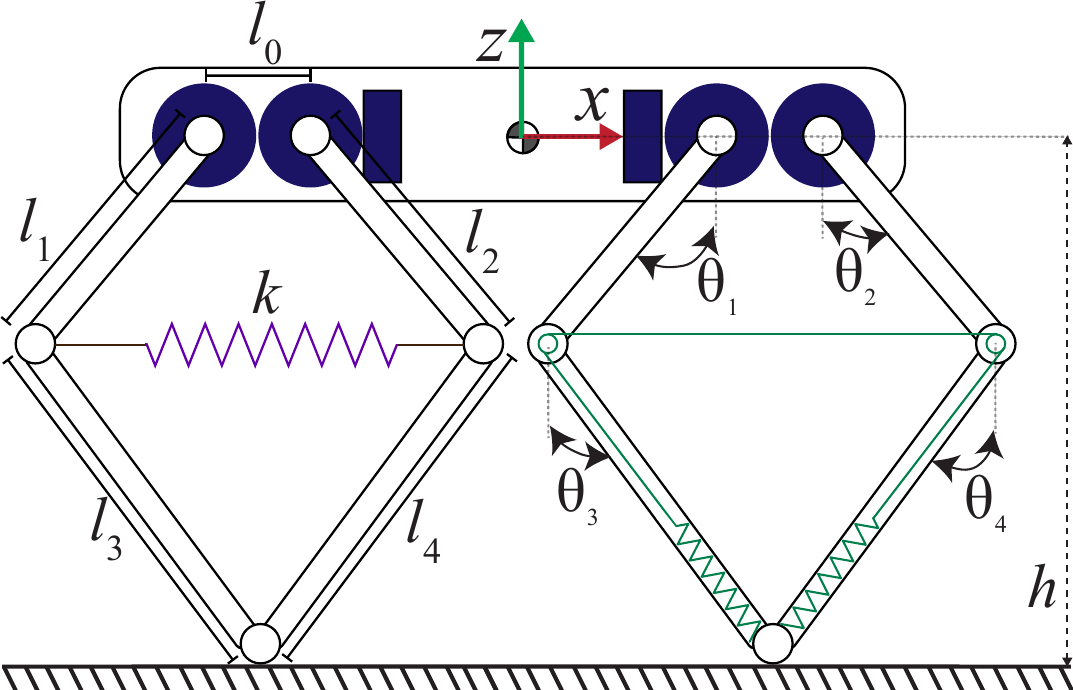}
    \vspace{-2ex}
    \caption{Illustration of the 5-bar mechanism on a quadruped, in side view. The right leg represents the proposed spring pulley integration, whereas the left leg displays the simplified spring used in the simulation.}
    \label{fig:linkExplain}
\end{figure}

Furthermore, the 5-bar mechanism can incorporate springs to increase the jumping capabilities utilizing a pulley system in the lower links. These springs are connected through a cord from the knees and located near the paw, which helps increase inertia for stabilization during flight, as illustrated in Fig.~\ref{fig:linkExplain} with one actuator actuating the hips and two actuating the 5-bar mechanism for standing, walking, jumping, and in-flight attitude changes.  When paired with a string, the mechanism permits two energy-efficient resting postures: one initiated when the springs engage during standing, and the other in a squat position, where the spring force aligns directly with the motor, preventing leg actuation.

\subsection{Actuation, electronics, and sensing} 

To actuate the legs for jumping, standing, walking, and in-flight attitude control, a high torque actuator was selected, and its model is used in all simulation studies presented in this work. Specifically, the actuator is a relatively low-cost, commercial, off-the-shelf motor with a high torque output and torque density, the CubeMars AK70-10, with the V2.1 driver board controlled over CAN-bus. It delivers a maximum peak output torque of $24.8\ \textrm{Nm}$.

An Asus Next Unit of Computing (NUC) with an AMD Ryzen 7 5700U CPU is selected as the onboard computer to control the motors through a USB to CAN-bus adapter and offer computational resources for autonomy. 

Regarding sensors, the robot integrates a VectorNav VN-100 Inertial Measurement Unit (IMU) and four Intel RealSense D455 Depth Cameras positioned at the front, back, left, and right sides. The D455 sensor has a depth field of view equal to $87\times 58$ degrees. The front and back cameras are angled down by $10$ degrees, while the left and right cameras are angled at $60$ degrees to capture images of the ground directly below and to either side of the robot. These cameras provide stereoscopic imaging to create the robots' local and global maps of the lava tube. The direct depth sensing from the sensors is critical for the robot's locomotion, while through IMU fusion of the additional RGB camera of the D455 for visual-inertial odometry~\cite{bloesch2015robust} --as used in our earlier DARPA Subterranean Challenge work~\cite{tranzatto_cerberus_2022}-- maps can be constructed, which are essential for local navigation. The robot is equipped with four LEDs positioned in the front and back to illuminate its immediate environment, enhancing both navigational precision and image acquisition quality.

Powering its array of actuators, sensors, and auxiliary electronics is a 4-piece battery pack designed to provide large bursts of current for powerful jumps while enabling an extended operational duration. The current design utilizes a  $44.4 \ \textrm{V}$ battery with a capacity of $2400 \ \textrm{mAh}$ and a current delivery capacity of $ 150 \ \textrm{C}$ weighing $0.7 \textrm{kg}$. The power electronics have been dimensioned based on the peak energy demands inherent in the jumping phase. This ensures they are aptly suited for high-power jumping and sustained pronking, a form of burst energy locomotion. Simultaneously, these electronics support fundamental actuation operations such as walking and standing. In these more stationary phases, the robot actively maps its surroundings, formulating efficient traversal routes, and determining jump trajectories to overcome obstacles.

The robot has been designed to accommodate a payload capacity of up to $2.5\ \textrm{kg}$. Notably, an increase in ferried payload directly influences and proportionally diminishes the robot's jumping efficacy. As a representative setup for baseline evaluations, a lightweight Raman spectrometer, weighing $0.25\ \textrm{kg}$, is mounted to the robot's forward-facing region, serving as the model payload for simulations. Several other possible payloads could be used for this mission. Examples with similar mass include: 1. A front-facing or $360 \ \textrm{deg}$ LIDAR, which could be used to create a higher quality 3D map of the lava tube (e.g., Livox Mid-360). 2. A high-quality infrared or low-light camera with a more powerful, direct light source (e.g., FLIR Tau 2). 3. Higher power communication equipment useful for coordination in case of deployment with other robots in a swarm (e.g., Rajant DX2). 4. Other possible payloads could be considered.

\subsection{Link length grid search optimization}

A systematic analysis of different link lengths and spring stiffness was conducted to tailor the 5-bar mechanism for optimal jumping performance and appropriately dimension the robot. This evaluation aimed at ascertaining the combination of parameters that could maximize the jump height under simulated Martian gravity conditions.

The simulation environment utilized was MATLAB's Simscape Multibody. Within this framework, an exhaustive grid search was performed by iteratively modifying the link lengths and spring stiffness. The performance of each configuration was evaluated based on the height of a simulated jump, with predefined squat angles and torque limits. The Simscape simulations utilized a variable solver switching between \texttt{ode15}, \texttt{ode23}, and \texttt{ode45} to ensure adequate accuracy and fidelity of the simulation output. The motors were simulated using actuated rotational joints with state feedback into a PID controller tracking the reference position for the squat and jump, which outputs torque commands to the motors. The rotational joints representing the motor model were tuned with damping coefficients and torque saturation limits to replicate behavior seen in physical tests of the motors regarding output response and peak rotational velocity. Knees and ankles were represented with passive rotational joints.

In the simulation framework, the robot's main body was represented by a cuboid encapsulating its dimensions, weight, and inertia properties. The legs were represented by links with the density of aluminum for the hip links $l_1$ and $l_2$, and carbon tube for links $l_3$, $l_4$ to account for the incremental weight effects of elongating links. For the simulation, the rotation and translation of the body were constrained to a one-dimensional movement vertically. At the same time, the simulated actuators were given setpoints to the first squat and then perform the jump. The squat and jump sequence began with motor setpoints at $17 \ \text{deg}$. Over $1.5 \ \text{s}$, this setpoint increased linearly to \(120 \, \text{deg}\), after which it returned to $17 \ \text{deg}$ in a single timestep. The PID controller generated the corresponding torques for the jump. Torque saturation was limited to $20.32 \ \text{Nm}$ during the jump tests, incorporating a safety margin of $10\ \% $. $17 \ \text{deg}$ selection was based on the spring's natural length in simulation and the safety margin it afforded, preventing knee collisions during motor overshoot. The parameters altered for each simulation iteration were $l_1$, $l_2$, $l_3$, $l_4$, and spring stiffness $k$. The range of values simulated can be seen in Table~\ref{tab:ranges_sim}. The link $l_0$ is set to $0.09 \  \textrm{m}$ as it is the selected distance between the motors based on the minimal distance of the axles of the motors when placed side by side.

\begin{table}[t!]
    \centering
    \caption{Grid search optimization simulations parameters.}
    \vspace{-2ex}
    \resizebox{0.45\textwidth}{!}
    {\begin{tabular}{llr} \Xhline{3\arrayrulewidth}
    \textbf{Parameter } & \textbf{Symbol} & \textbf{Value}\\\hline

    Mass of single leg     & $m_l$   & $2.0$ {[}kg{]}       \\
    Mass of payload        & $m_b$   & $0.25$  {[}kg{]} \\
    Body mass              & $m_{body} $ &  $6.55$  {[}kg{]} \\
    Total quadrupedal robot mass   & $m_{robot}$   & $14.8$ {[}kg{]}       \\
    Link 0 length          & $l_0$   & $0.09$ \ {[}m{]}  \\
    Link 1 length range    & $l_1$   & $0.10-0.35$ \ {[}m{]}  \\
    Link 2 length range    & $l_2$   & $0.10-0.35$ \ {[}m{]}  \\
    Link 3 length range    & $l_3$   & $0.15-0.45$ \ {[}m{]}  \\
    Link 4 length range    & $l_4$   & $0.15-0.45$ \ {[}m{]}  \\
    Spring stiffness       & $k$     & $600-1000$ {[}N/m{]}      \\ \Xhline{3\arrayrulewidth}
    \end{tabular}}
    \label{tab:ranges_sim}
\end{table}

\begin{figure}[t]
    \centering
    \includegraphics[width =0.46\textwidth]{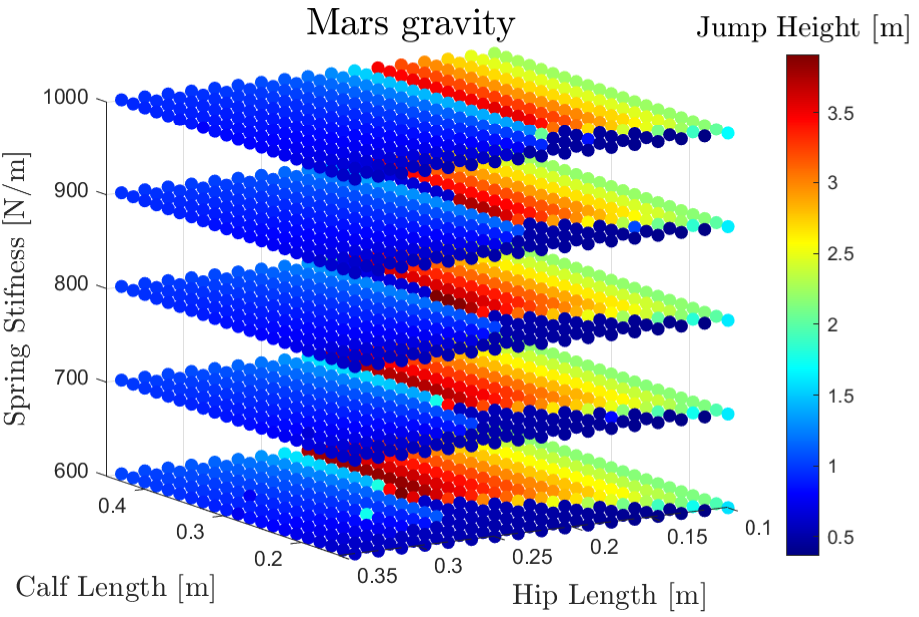}
    \vspace{-2ex}
    \caption{4D representation for the jump heights reached in simplified 4-legged robot simulations, using Mars gravity.}
    \label{fig:gridSearch}
\end{figure}

Fig.~\ref{fig:gridSearch} visualizes the array of simulated jump heights, measured from the center of mass,  corresponding to the varied parameters. The link lengths were selected based on the achieved jump height while considering design constraints like size-weight relations for the entire robot and appropriate control authority margins for motor actuation with integrated springs. It is important to secure adequate torque margins for the motor to actuate the spring effectively, preventing the necessity of utilizing its full torque output for spring extension. Choosing a hip length, calf length, and spring stiffness is a compromise between achieving a maximum jump height and having control authority margins for the motors, ensuring the ability to walk and proficiently execute smaller jumps efficiently. A shorter hip and calf length than the length for optimal jump height was deemed beneficial as it facilitated the control margins and a more compact arrangement of legs on the robot chassis, which reduced the robot's overall weight by constraining its size, which directly affects jumping capabilities. 

These design considerations reduced the robot's overall size, minimized weight, and enhanced its structural rigidity. The leg parameters selected for the design are listed along the optimal parameters based on jump height in Table~\ref{tab:selcectedParams}.

\begin{table}[t]
    \centering
    \caption{Chosen and optimal leg parameters for OLYMPUS }
        \vspace{-2ex}
    \resizebox{0.43\textwidth}{!}
    {\begin{tabular}{lccc} \Xhline{3\arrayrulewidth}
    \textbf{Parameter } & \textbf{Symbol} & \textbf{Selected} & \textbf{Optimal}\\\hline
    Link 0 length      & $l_0$   & $0.09$ \ {[}m{]}  & $0.09$ \ {[}m{]}  \\
    Link 1 length      & $l_1$   & $0.175$ \ {[}m{]} & $0.205$ \ {[}m{]} \\
    Link 2 length      & $l_2$   & $0.175$ \ {[}m{]} & $0.205$ \ {[}m{]} \\
    Link 3 length      & $l_3$   & $0.3$ \ {[}m{]}   & $0.315$ \ {[}m{]} \\
    Link 4 length      & $l_4$   & $0.3$ \ {[}m{]}   & $0.315$ \ {[}m{]} \\
    Spring stiffness   & $k$     & $800$ {[}N/m{]}   & $800$ {[}N/m{]}   \\  
    Simulated jump height         & $h$     & $3.4$  \ {[}m{]}  & $3.9$  \ {[}m{]}\\ \Xhline{3\arrayrulewidth}
    \end{tabular}}
    \label{tab:selcectedParams}
\end{table}

A notable decrease in maximum jump height was observed between the optimal jump height leg parameters and the chosen leg parameters, but the increased jump versatility was considered a justified trade-off.

\subsection{Quadrupedal robot design}
\begin{figure}[t]
    \centering
    \includegraphics[width = 0.49\textwidth]{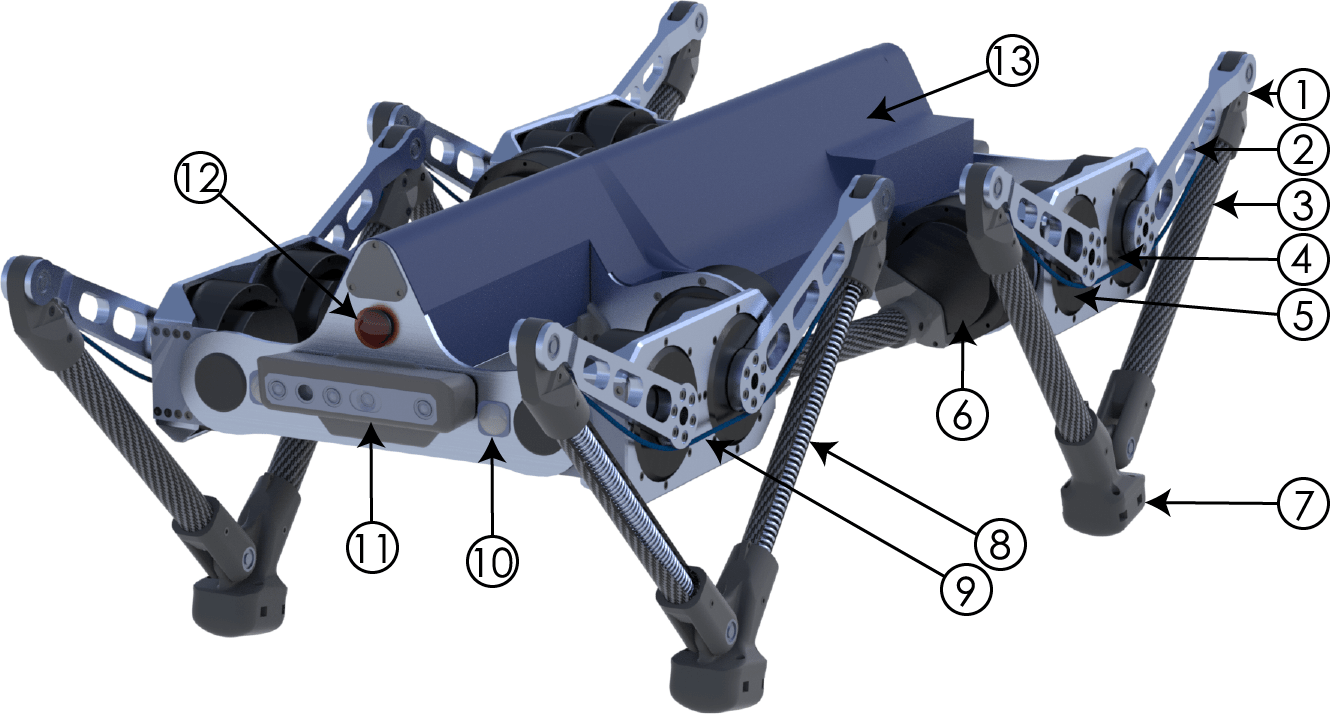}
    \vspace{-4ex}
    \caption{Rendered view of OLYMPUS design, in a squat position with a red payloads lens mounted facing forward.}
    \label{fig:CADFigure}
\end{figure}

The final designed 5-bar leg optimized for jump height based on the selected link lengths and spring stiffness is depicted in Fig.~\ref{fig:CADFigure}.

The OLYMPUS quadruped consists of four 5-bar legs attached to a body. The distance between the two front legs was set based on their ability to rotate freely $180\ \textrm{deg}$ up without touching the body and then adding the width of the structural beam of the robot spine. The hip actuators were placed towards the center of the robot to lessen rotational inertia. The distance between the back legs was set to be $5\ \textrm{cm}$ wider on each side of the robot to enable the back legs to have a full workspace without the risk of hitting the front legs, as the front legs need some limits in their movements for some configurations to eliminate the chances of the legs colliding. The length-wise separation of the front and back legs was set to the minimum possible distance while still having the distance long enough to remove the possibility of the knees colliding while squatting or performing in-flight attitude corrections.

An image of the quadrupedal jumping-legged robot can be seen in Fig.~\ref{fig:CADFigure}, with numbered annotations as follows: 1. Knee, 2. Hip, 3. Carbon Tube Calf, 4. 5-bar Motor, 5. 5-bar Motor, 6. Hip Motor, 7. Paw, 8. Integrated Spring (fully extended in squat pose), 9. High-Strength Cord, 10. LED, 11. Depth and RGB Camera, 12. Payload Lens, 13. Body (comprising compute unit, IMU, battery, and payload). The front left leg is shown with a section view to make the integrated springs visible.

\vspace{-2ex}
\section{Simulation}\label{sec:070_Simulation}
This section describes simulations evaluating OLYMPUS' motion properties, including locomotion, jumping, landing, and in-flight maneuvers. These simulations were performed with the final parameters and design from Fig.~\ref{fig:CADFigure} in MATLAB Simscape, with Martian gravity of $3.71 \frac{m}{s^2}2$.

\subsection{Locomotion}
Traversing relatively flat expanses of Martian terrain on the surface and sections within Martian lava tubes that may be traversable without significant jumps necessitates specific locomotion strategies to optimize energy consumption and adaptability. Three locomotion methods stand out for these terrains:  walking, pronking, and sequential smaller jumps. Walking employs a consistent movement pattern, efficiently utilizing the robot's mechanisms for continuous travel over longer distances.

The main forms of walking for quadrupedal robots are bounding, pronking, pacing, trotting, and dynamic walking \cite{papadopoulos2013ariadna}. Pronking, characterized by a series of coordinated jumps using all four legs simultaneously, offers rapid movement over shorter distances or when terrain contains small obstacles that need to be traversed. Pronking is a locomotion strategy that benefits from lower gravity \cite{kolvenbach2018scalability}. 

By utilizing pronking or a series of smaller jumps, OLYMPUS can reach a forward velocity of $0.51\ \frac{m}{s}$ with a very basic locomotion gate and easily clear small boulders or small cracks up to $0.5\ \textrm{m}$ in size when utilizing pronking or small jumps, which could be an expected rock size on the Martian surface~\cite{golombek2003rock}.

\subsection{Jumping and landing}

\begin{figure}[t]
    \centering
    \includegraphics[width = 0.49\textwidth]{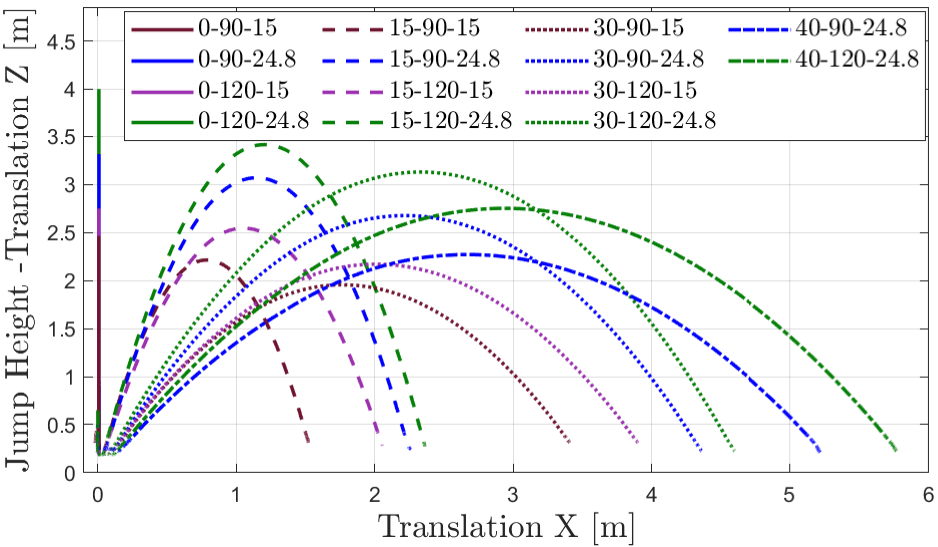}
    \vspace{-2ex}
    \caption{Simulated jumps with jump parameters from Table~\ref{tab:jumpSimParams}.}
    \label{fig:JumpTestsSim}
\end{figure}

The robot's jumping capabilities are the primary focus of the design. Additional simulation studies were conducted to assess its ability to use jumping as a form of locomotion and overcome significant obstacles. A series of jumps with varying motor torque, squat angle, and jump angle were performed to determine OLYMPUS' jumping capabilities in simulation. Table~\ref{tab:jumpSimParams} shows the time in flight ($t$), height at apogee ($h_{max}$), and maximum horizontal distance ($x_{max}$) for the different jumps, while their corresponding trajectories are presented in Fig.~\ref{fig:JumpTestsSim}, all in Mars gravity. The height is measured from the center of mass.

The squat angle pertains to the 5-bar motor angle used for the jump, while the jump angle denotes the orientation of a vector from the paw to the midpoint of $l_0$  for each leg and its angle relative to the vertical z-axis in the global frame.

The OLYMPUS robot can, when jumping vertically, reach a height of $4.01\ \textrm{m}$ and has a maximum flight time of $2.72\ \textrm{s}$ in simulation when using the max actuator output of $24.8\ \textrm{Nm}$ and maximum squat angle of $120\ \textrm{deg}$. In simulation jumps with $40^\circ$ jump angle and $90^\circ$ or $120^\circ$, with $\tau_{out}$ = $15\ \textrm{Nm}$, paw slippage was observed, rendering the jump invalid. Landings were performed with timed squats to dampen the velocity from the jump to zero.

\begin{table}[t]
    \centering
    \caption{Simulated jump trajectory data.}
    \vspace{-2ex}
    \resizebox{0.47\textwidth}{!}
    {\begin{tabular}{lccccccc} \Xhline{3\arrayrulewidth}
        \textbf{Jump} & \textbf{Squat} & \multicolumn{3}{c}{$\tau_{out}$ = $15$ Nm} & \multicolumn{3}{c}{$\tau_{out}$ = $24.8$ Nm} \\
    \textbf{Angle}  & \textbf{Angle}     & $h_{max}$ & $x_{max}$ & $t$ & $h_{max}$ & $x_{max}$ & $t$ \\ \hline
    $0^\circ$     & $90^\circ$   &   $2.47$ &  $0$  & $2.1$   & $3.33$   & $0$   & $2.22$   \\
    $0^\circ$     & $120^\circ$  &   $2.75$ &   $0$ &  $2.25$  &  $4.01$  &  $0$  & $2.72$   \\
    $15^\circ$    & $90^\circ$   &  $2.21$  &  $1.63$  &  $1.9$  &  $3.1$  &  $2.36$  &  $2.4$  \\
    $15^\circ$    & $120^\circ$  &   $2.55$ &  $2.08$  &  $2.8$  &  $2.41$  & $2.39$   &  $2.51$  \\
    $30^\circ$    & $90^\circ$   &   $1.96$ &  $2.52$  &  $1.8$  &  $2.68$  & $4.48$  &  $2.22$  \\
    $30^\circ$    & $120^\circ$  &  $2.17$  &  $3.94$  &  $1.92$  &  $3.13$  & $ 5.77$  & $2.37$   \\
    $40^\circ$    & $90^\circ$   &  -  &  -  &  -  &  $2.27$  &  $5.43$  &  $1.95$  \\
    $40^\circ$    & $120^\circ$  &  -  &  -  &  -  &  $2.75$  &  $5.94$  &  $2.2$  \\ \Xhline{3\arrayrulewidth}
    \end{tabular}}
    \label{tab:jumpSimParams}
\end{table}

From the data presented in Table~\ref{tab:jumpSimParams} and the trajectories shown in Fig.~\ref{fig:JumpTestsSim}, given a vertical distance of $0.15\ \textrm{m}$ bottom of the robot and the Center of Mass(COM), the calculated maximum ground clearance with legs fully squatted during flight is $3.86\ \textrm{m}$. Simulations to measure the effect of larger payloads demonstrated that for every $1$ kg increase in weight, jump height decreases by $0.15\ \textrm{m, } 0.14\ \textrm{m}$, and $0.13$ m for the first, second, and third kg added to the simulated mass.

\subsection{In-flight attitude corrections}

The robot's flight duration post-takeoff and prior to landing is governed by the vertical velocity component during jumps. Consequently, an increased vertical velocity results in an extended flight time. It should be noted that this assertion is based on the assumption of flat ground conditions. To execute powerful jumps, the robot's legs must be synchronized accurately in terms of timing, torque profile, and motor angles. Minor deviations in motor angles, inaccuracies in torque profiles, or variations in paw liftoff timing will influence the robot's rotational velocity, altering its jump trajectory and rotation. Simulations indicate that appropriate alignment of the hip motors and torque profiles mainly impacts the pitch axis. In contrast, the roll is affected mostly by the differential actuation of the left or right legs. Any offset in hip motors predominantly drives yaw rotation, causing the robot to revolve about its vertical axis. 

Simulations demonstrate that even high-powered successful forward jumps can result in angular velocities for roll up to  $8\ ^{\circ} \textrm{/s}$ in normal jumps, $30\ ^{\circ} \textrm{/s}$ for pitch, and $5\ ^{\circ} \textrm{/s}$ for yaw. Angular rates depend on the jump's type and power, e.g., vertical, horizontal, and sideways. The tendency of higher angular rates for pitch is primarily attributable to the torque generated by leg forces stemming from an unequal lever arm relative to the COM during angled jumps, while the moment around roll and yaw are mostly symmetric for a forward jump from a flat surface. 

This underscores the importance of in-flight stabilization to enable the robot to land safely. Adjusting the robot's mid-flight orientation is important for safety and preventing damage upon landing. For OLYMPUS, in-flight stabilization maneuvers can be facilitated by the robot moving its legs and thus experiencing reaction torques on the main torso due to the mass/inertia distribution variation due to the leg motion. 

\begin{figure}[t]
    \centering
    \includegraphics[width = 0.49\textwidth]{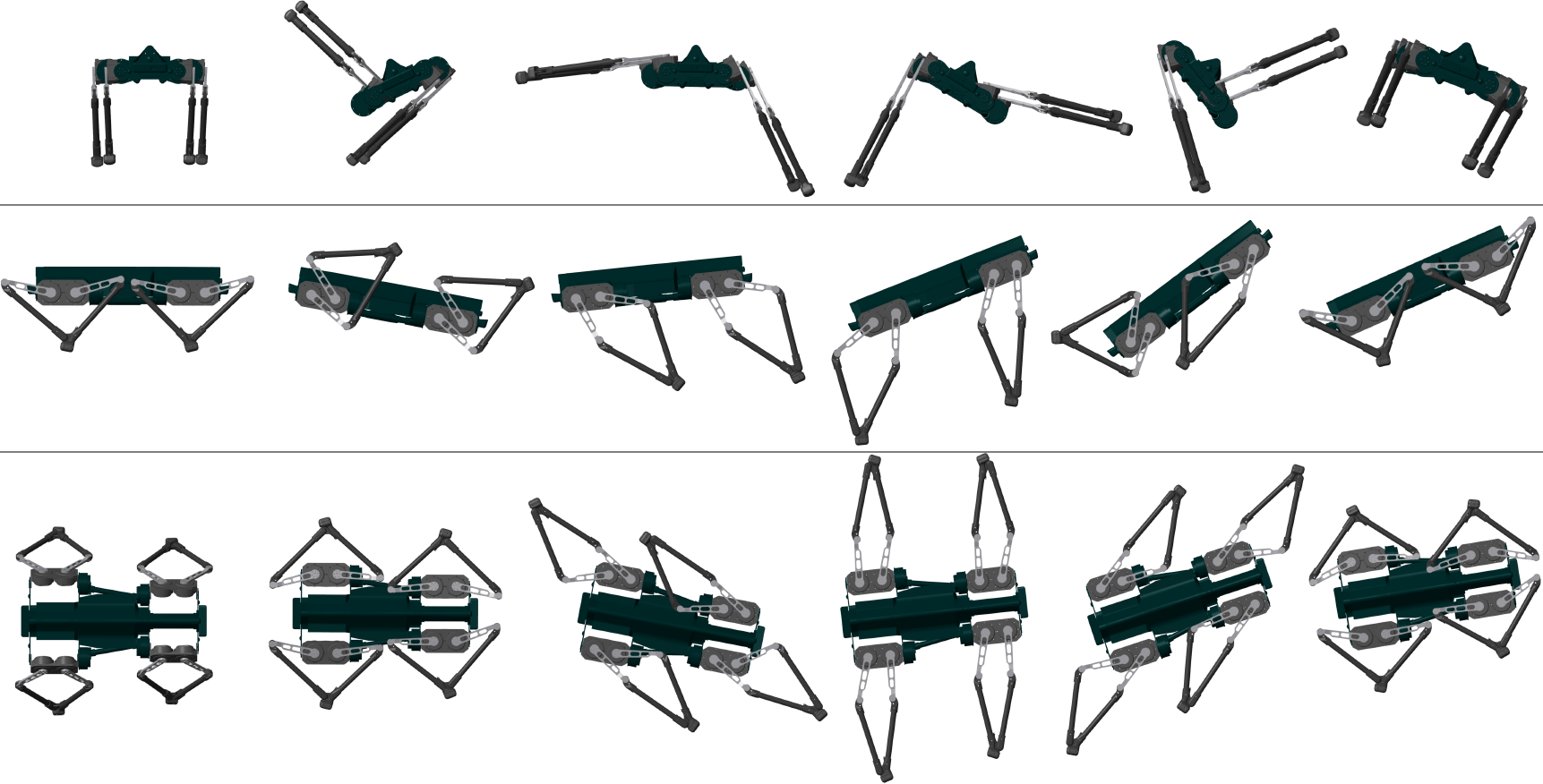}
    \vspace{-4ex}
    \caption{Image sequence of the sequential corrections in roll, pitch, and yaw angles. The sequence progresses from left to right within one movement cycle. The final frame demonstrates the angular adjustment post-cycle. Roll is depicted at the top, pitch in the center, and yaw at the bottom.}
    \label{fig:roll_pich_yaw_composit}
\end{figure}

To determine the ability of a robot to correct any angular velocity given during a jump, a series of coordinated moves were simulated specifically for roll, pitch, and yaw rotations in flight. Fig.~\ref{fig:roll_pich_yaw_composit} illustrates roll, pitch, and yaw moves. Examples of the robots' reachable leg configurations, workspace, and positions are also illustrated.

\setlength{\belowcaptionskip}{-10pt}  
\setlength{\belowcaptionskip}{0pt}  

\begin{figure}[t]
    \centering
    \includegraphics[width = 0.49\textwidth]{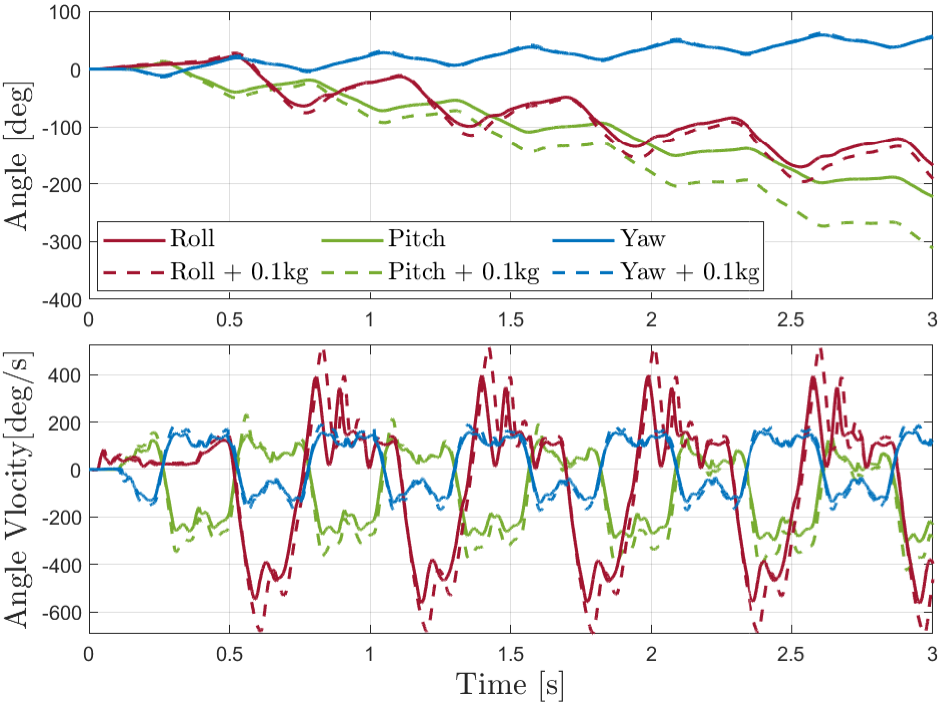}
    \vspace{-2ex}
    \caption{Roll, pitch, and yaw angles, coupled with their respective angular velocities, during single-axis in-flight correction maneuvers.}
    \label{fig:roll_pich_yaw_3sec}
\end{figure}

These coordinated moves were simulated to measure the maximum angular velocity and angle the robot can correct during a $3\ \textrm{s}$ window, one axis at a time. The simulations were conducted on a free-floating robot. The resulting angular velocity and angles achieved in $3$ seconds are illustrated in Fig.~\ref{fig:roll_pich_yaw_3sec}. The simulation employing a $0.1\ \textrm{kg}$ point mass demonstrates the augmented angular velocity resulting from increased inertia at the paw.

The resultant angular velocity reached, averaged over $3 \ \textrm{s}$, is $73.3 \ \textrm{$^{\circ}$/s}$ in roll, $55.1 \ \textrm{$^{\circ}$/s}$ in pitch and $18.3 \ \textrm{$^{\circ}$/s}$ in yaw.

\vspace{-2ex}
\section{Mission Profile} \label{sec:080_MissionProfile}
This section outlines a proposed mission for the presented jumping quadrupedal robot to explore a lava tube on Mars.
\begin{figure*}[t]
    \centering
    \includegraphics[width = 0.995\textwidth]{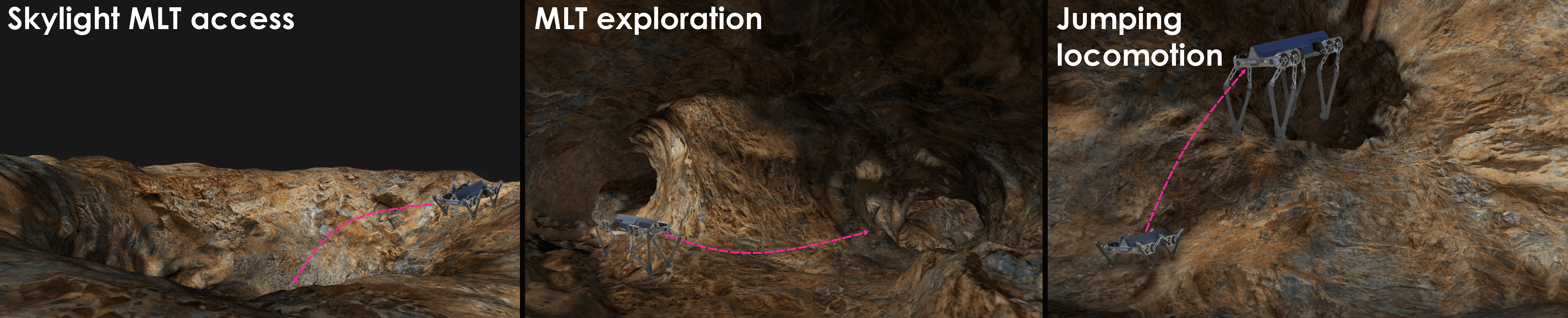}
    \caption{Illustration of OLYMPUS used in proposed Martian lava Tube (MLT) exploration. The visualized size of the robot is magnified to allow its presentation in the otherwise vast expected size of MLTs.}
    \label{fig:SynFigure}
\end{figure*}

\subsection{Mission objective}
The mission objectives behind a Martian lava tube exploration mission are numerous. Firstly, exploring these lava tubes offers unparalleled insights into Mars's geological evolution, serving as a rich repository of its past volcanic history. Secondly, these underground formations are potential candidates for future human habitation. A comprehensive exploration would ascertain the availability of crucial resources like water-ice and valuable minerals, vital for sustaining long-term human presence. Moreover, detailed measurements of temperature variations, radiation levels, and the capability of these tubes to provide environmental shielding are essential parameters to assess. 

The proposed mission has delineated a series of critical objectives to ensure a comprehensive exploration of the Martian lava tubes. These objectives are as follows: 1) Ingress and Survey: Initiate entry into the lava tube and explore the initial $200-500\ \textrm{m}$ from the entrance, constrained by the battery life and the unknown difficulty of traversability~\cite{mari2021potential}. This phase will employ sensing to generate a high-fidelity 3D topographical map while also capturing detailed images of the lava tube interiors. 2) Geochemical Analysis: Deploy a spectrometer to decipher the elemental and molecular composition of the tube's terrain and walls. The robot identifies data collection points based on predefined criteria, ensuring a comprehensive and representative sample set. 3) Data Transmission and Sustenance: Once the primary exploration phase concludes, the robot will navigate back to the lava tube entrance to transmit its captured data and, if possible, recharge for continued operation.

To perform this task, we propose a team of three OLYMPUS robots utilizing a similar strategy as~\cite{arm2023scientific} with the three robots having different payloads based on their task. For payload configurations: OLYMPUS 1 is outfitted with a camera and LiDAR; OLYMPUS 2 incorporates a LiDAR and a lightweight spectrometer; and OLYMPUS 3 is equipped solely with a spectrometer.

\vspace{-1ex}
\subsection{Lava tube selection and mission start}
For the proposed mission, we've selected a region within the Tharsis area of Mars with a high density of lava tubes~\cite{mari2021potential}. A rille linked to a $50$-meter diameter lava tube, inferred from its collapsed segments, was chosen due to its closeness to a suitable landing zone characterized by relatively flat terrain.

After deployment from the lander, situated approximately $500$ meters from the targeted lava tube \cite{miaja2022robocrane}, the unit, consisting of three OLYMPUS robots and an auxiliary rover with a comprehensive solar panel array and antenna for energy and data relay, traverses to the rille entrance across flat Martian terrain. Upon reaching the destination, each robot initiates battery replenishment to optimize operational duration within the lava tube.

\subsection{Entrance of lava tube}
The three robots begin entering the lava tube by first mapping the rille opening and determining the best path to enter the cave. They use high-resolution imagery from the MRO satellite to aid in their decision. The cave entrance is defined by a large amount of fallen rocks and rock piles. The robots traverse the initial section from the rover to the overhanging section of the cave entrance by walking and making small jumps, clearing $0.5\  \textrm{m}$-sized rocks~\cite{golombek2003rock}. As they close in on the start of the overhanging section of the cave, the robots perform a single powerful vertical jump to more effectively see the terrain ahead and create a map for the first section of the entrance. By utilizing the jump and scanning technique and sharing their individual LiDAR scans, the robots are able to create a collective map of the entrance, which they relay to the rover, lander, and mission control on Earth. Mission control then gives the go signal for the robots to enter the cave. The path is planned by the robots using frameworks such as the graph-based planning framework developed by~\cite{dang_graph-based_2020}, while localization and mapping are achieved through methods such as~\cite{khattak2020complementary}. Essentially, the core autonomy functionality may be built on top of prior work on the high-relevant DARPA Subterranean Challenge~\cite{tranzatto_cerberus_2022}. The path selected is initially blocked by roof collapse debris~\cite{mari2021potential}. First, a $2$ meter-high and $2$ meter-deep rock pile, and a few meters further in, a $2.5$ meter boulder blocks the path. These boulders and rouble piles are estimated to be $5-10$ meters in size on the moon~\cite{sauro_lava_2020}; thus, a rock size decrease relative to the decrease in lava tube dimensions could be expected on Mars. The OLYMPUS robots are able to jump over these obstacles. During the first obstacle, one of the OLYMPUS robots lands and dampens all the velocity from the jump but falls to one side due to a conservative leg angle. It performs a self-right maneuver and stands up without significant damage. It then relays the information to the other robots, which land safely after the first jump. While performing the last jump, two of the robots experience an unwanted pitch spin of $10$ degrees per second, which they are able to correct using in-flight maneuvers to ensure a safe landing. The robots are then past the theorized section of significant rubble at the rille entrance.

\subsection{Exploration and science inside the lava tube}
The robots make additional data collection points with cameras and the spectrometer before going into the cave and also collect data points while traversing the collapsed section of the rille entrance. These readings are transmitted to the rover as they are collected before the communication with the rover is blocked by the rubble pile after the trio of robots has entered the cave. Once clear of the entrance rubble pile, the robots autonomously map and select a suitable number of sample points of interest based on predefined conditions, and the robots autonomously decide where to collect data points while also annotating in their shared map where the samples were taken. The annotation takes place using a volumetric dense representation of the environment such as voxblox~\cite{oleynikova2017voxblox} much like in our previous work on the DARPA Subterranean Challenge~\cite{tranzatto_cerberus_2022}.

\subsection{Exit from the lava tube/transmission of data at end of mission}
Once either the robots have reached a predetermined battery percentage, a preset time limit has been reached, or they have collected samples and mapped the first accessible section of the cave, they return to the rubble filled rille entrance. They then transmit the collected data to the rover, which forwards it to the lander and finally to mission control. The scientists then use this information to determine the structure, size, and composition of the lava tube. Fig.~\ref{fig:SynFigure} depicts the OLYMPUS robot during its lava tube exploration mission. The robots can also drop communication breadcrumbs while entering the lava tube to ensure communication and transfer of data for more extensive exploration missions.

\section{Conclusions and further work}\label{sec:conclusions}

Lava tube exploration is motivated by scientific research, locating resources, and potential human exploration, while at the same time presenting significant challenges. This paper assesses traditional exploration techniques and presents novel robotic and legged robotic solutions for the task.

We further present OLYMPUS, a jumping legged quadruped for Martian lava tube exploration. Utilizing fast high torque motors in conjunction with incorporated springs to optimize jump height, the system achieves a simulated jump height of $4.01 \textrm{m}$ in Martian gravity. The utilization of a 5-bar mechanism in its legs, with optimized link lengths and spring stiffness, allows for efficient walking and optimized jumping. Furthermore, the robot is equipped with the ability to stabilize its rotational velocity and correct its attitude mid-flight. Capable of correcting $73.3 \ \textrm{deg/s}$ in roll, $55.1 \ \textrm{deg/s}$ in pitch, and $18.3 \ \textrm{deg/s}$ in yaw. These results underline the utility of OLYMPUS for Martian lava tube exploration. 

A mission concept involving a trio of OLYMPUS robots for Martian lava tube exploration was also presented. Subsequent research will involve prototyping and assessments of the robot. Additionally, rigorous evaluations of varying terrains like regolith, sand, rubble, and rock will assess the robot's adaptability. Advanced control methodologies for robust locomotion, including walking, pronking, jumping, landing, and mid-flight attitude adjustments, are also to be developed.


\printbibliography

\end{document}